\documentclass[10pt,twocolumn,letterpaper]{article}

\usepackage{iccv}              

\definecolor{iccvblue}{rgb}{0.21,0.49,0.74}
\usepackage[pagebackref,breaklinks,colorlinks,allcolors=iccvblue]{hyperref}

\def\paperID{51} 
\def\confName{Third Workshop on Computer Vision for Automated Medical Diagnosis}
\def\confYear{2025}

\title{Hierarchical Classification for Improved Histopathology Image Analysis}

\author{
Keunho Byeon$^{1}$ \quad 
Jinsol Song$^{1}$ \quad 
Seong Min Hong$^{2}$ \quad 
Yosep Chong$^{2}$ \quad 
Jin Tae Kwak$^{1}$\\
$^{1}$School of Electrical Engineering, Korea University, Seoul, Korea \\
$^{2}$Department of Hospital Pathology, The Catholic University of Korea College of Medicine, Seoul, Korea \\
{\tt\small 
\{bkh5922, truetg, jkwak\}@korea.ac.kr, 
\{min7033, ychong\}@catholic.ac.kr}
}

\begin{document}
\maketitle

\begin{abstract}%
Whole-slide image analysis is essential for diagnostic tasks in pathology, yet existing deep learning methods primarily rely on flat classification, ignoring hierarchical relationships among class labels. In this study, we propose HiClass, a hierarchical classification framework for improved histopathology image analysis, that enhances both coarse-grained and fine-grained WSI classification. Built based upon a multiple instance learning approach, HiClass extends it by introducing bidirectional feature integration that facilitates information exchange between coarse-grained and fine-grained feature representations, effectively learning hierarchical features. Moreover, we introduce tailored loss functions, including hierarchical consistency loss, intra- and inter-class distance loss, and group-wise cross-entropy loss, to further optimize hierarchical learning. We assess the performance of HiClass on a gastric biopsy dataset with 4 coarse-grained and 14 fine-grained classes, achieving superior classification performance for both coarse-grained classification and fine-grained classification. These results demonstrate the effectiveness of HiClass in improving WSI classification by capturing coarse-grained and fine-grained histopathological characteristics.
\end{abstract}

\begin{figure*}[htp]
\includegraphics[width=\textwidth]{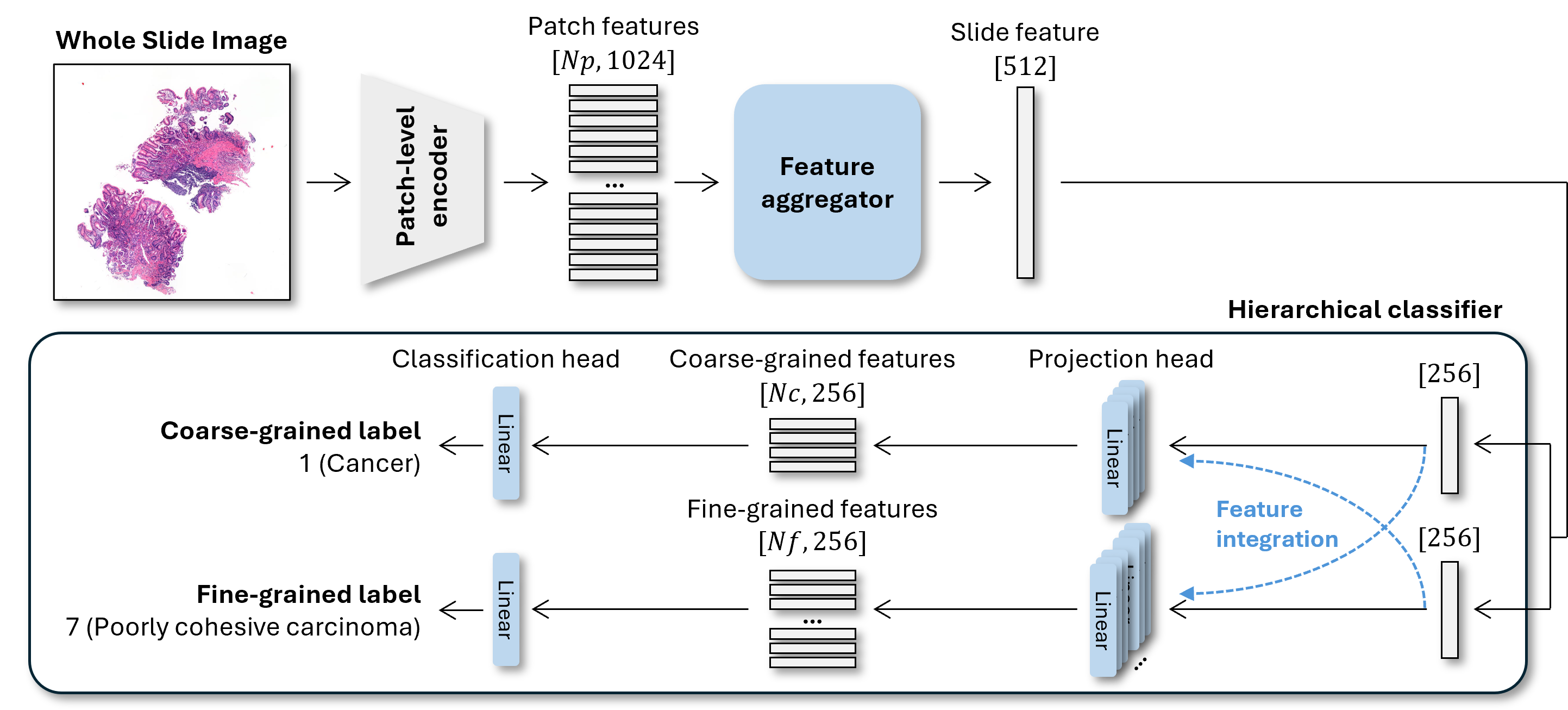}
\caption{
Overview of HiClass. \(N_p\), \(N_c\), and \(N_f\) represent the number of patches, coarse-grained classes, and fine-grained classes, respectively.
} 
\label{fig1}
\end{figure*}

\section{Introduction}
Whole-slide image (WSI) analysis plays a critical role in the diagnosis of various diseases in pathology \cite{shmatko2022,nguyen2024}. The advent of deep learning-based methods has significantly improved the performance of automated analysis at the WSI level \cite{bui2024,bui2024mecformer,jin2024}. Due to the large size of WSIs, they are typically processed using multiple instance learning (MIL), which enables efficient feature extraction and aggregation for classification. This approach is mostly formulated as either a binary classification or multi-class classification problem, which is referred to as a flat classification problem \cite{nguyen2024camp}. However, disease diagnosis naturally follows a hierarchical structure, where class labels are organized into a class hierarchy \cite{silla2011}. For example, a WSI may first be classified as benign or tumor. If classified as tumor, it can be further sub-categorized into well-differentiated, moderately-differentiated, and poorly-differentiated tumors. Despite this natural hierarchical structure, to the best of our knowledge, existing methods for WSI analysis primarily focus on flat classification, largely ignoring the inherent class hierarchy of WSIs.

Hierarchical classification categorizes WSIs at multiple levels, such as coarse-grained and fine-grained levels. Coarse-grained classification may include broader categories, e.g., benign and tumor, while fine-grained classification may involve detailed sub-categories, such as tumor differentiation grades. 
Although coarse-grained classification often achieves high accuracy, fine-grained classification remains challenging due to the higher inter-class similarity among fine-grained categories than coarse-grained categories and lack of sufficient samples for fine-grained classes \cite{li2024}. 
Integrating hierarchical classification into WSI analysis can offer several advantages, as it provides a more structured and clinically relevant approach that closely aligns with the diagnostic process used by pathologists. 
A few studies have explored hierarchical classification for pathology image analysis. For instance, Jin et al. \cite{jin2024} proposed hierarchical multi-instance learning (HMIL) that conducts both coarse-grained and fine-grained classification. HMIL was evaluated on three cancer datasets, such as cervical cytology cancer, breast cancer, and prostate cancer datasets. Similarly, Cai et al. \cite{cai2024} developed a Transformer-based model for cervical cytology classification across three hierarchical levels.
These studies suggest the potential of hierarchical classification for improved pathology image analysis.

In this study, we propose HiClass, a $\mathbf{H}$ierarchical $\mathbf{Class}$ification framework for improved histopathology image analysis. By incorporating class hierarchy into the classification process, HiClass improves WSI classification performance for both coarse-grained classification and fine-grained classification. To achieve this, HiClass integrates a multiple instance learning (MIL) framework with bidirectional feature integration, facilitating the information exchange between coarse-grained and fine-grained feature representations. To further optimize hierarchical learning, we introduce tailored, hierarchy-aware loss functions, including hierarchical consistency loss, intra- and inter-class distance loss, and group-wise cross-entropy loss. We evaluate HiClass on a dataset of gastric endoscopic biopsy slides that includes 4 coarse-grained classes and 14 fine-grained classes. The results demonstrate that HiClass achieves high hierarchical classification performance, effectively capturing both coarse-grained  and fine-grained histopathological characteristics.

\begin{figure*}[htp]
\includegraphics[width=\textwidth]{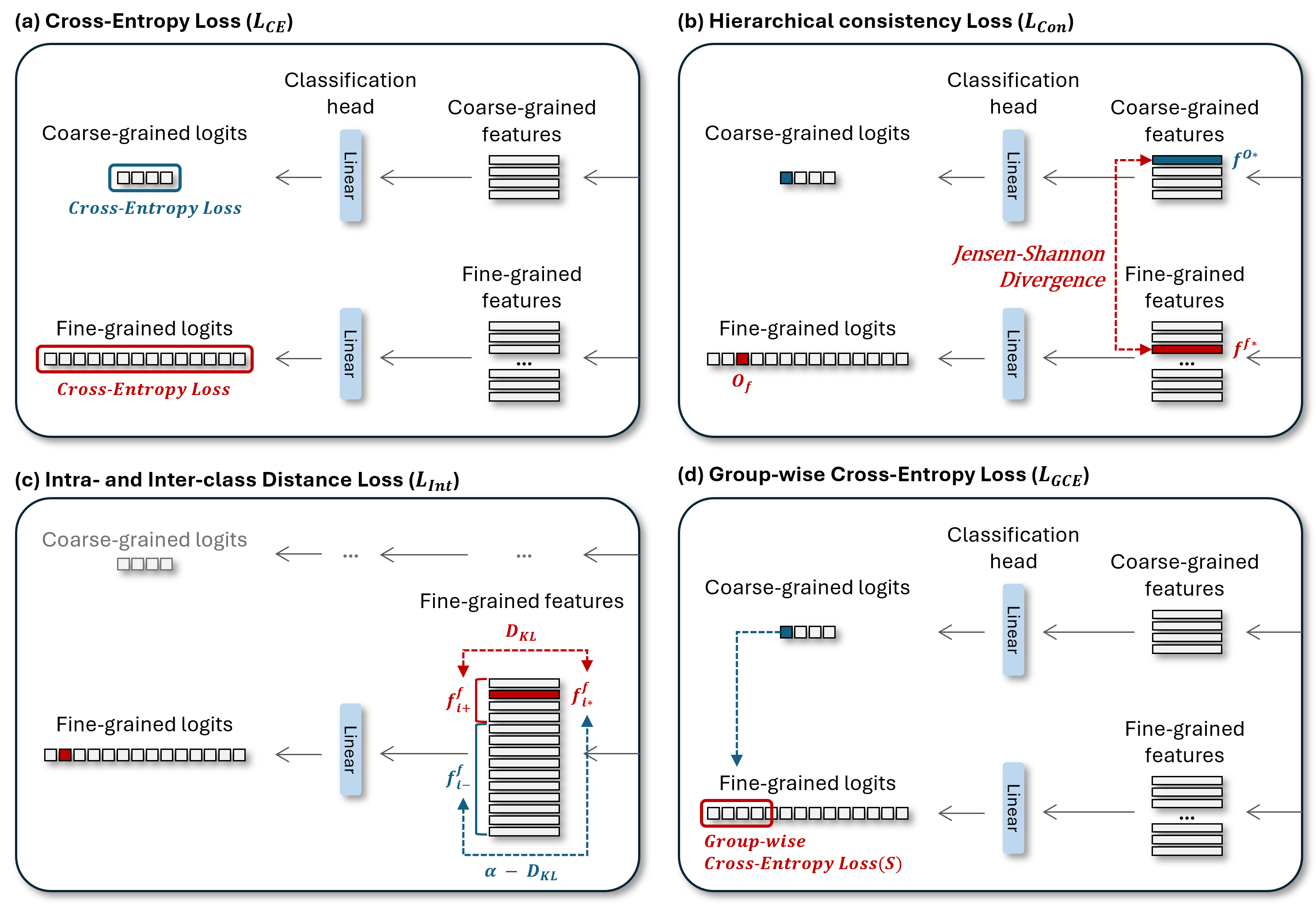}
\caption{
Illustration of proposed hierarchy-aware loss functions. (a) Standard cross-entropy loss is applied independently to coarse- and fine-level predictions. (b) Hierarchical consistency loss ($L_{Con}$) aligns the most confident coarse- and fine-level features using Jensen-Shannon divergence. (c) Intra- and inter-class distance loss ($L_{Int}$) encourages coarse-level grouping in the fine-grained feature space through margin-based KL divergence. (d) Group-wise cross-entropy loss ($L_{GCE}$) limits the fine-level prediction space to classes within the predicted coarse category, improving intra-group discrimination.
} 
\label{fig2}
\end{figure*}

\begin{table*}[t]
    \centering
    \caption{Details of a gastric endoscopic biopsy dataset.}
    \label{tab:class_distribution}
    \begin{tabular}{llcccc}
        \hline
        \textbf{Coarse-grained} & \textbf{Fine-grained} & \multicolumn{4}{c}{\textbf{Num slides}} \\
        \cline{3-6}
         & & \textbf{Train} & \textbf{Val} & \textbf{Test} & \textbf{Total} \\
        \hline
        Benign & Fundic gland polyp & 725 & 90 & 90 & 905  \\
         & Hyperplastic polyp & 288 & 36 & 36 & 360  \\
         & Inflammatory fibroid polyp & 259 & 32 & 32 & 323 \\
         & Xanthoma & 10 & 2 & 2 & 14 \\
        \hline
        Cancer & Malignant lymphoma & 18 & 2 & 2 & 22 \\
         & Neuroendocrine tumor & 9 & 2 & 2 & 13 \\
         & Tubular adenocarcinoma & 549 & 68 & 68 & 685 \\
         & Poorly cohesive carcinoma & 118 & 14 & 14 & 146 \\
        \hline
        Dysplasia & Tubular adenoma (low-grade) & 671 & 83 & 83 & 837 \\
         & Tubular adenoma (high-grade) & 96 & 11 & 11 & 118 \\
        \hline
        Gastritis & Chronic active gastritis & 126 & 15 & 15 & 156 \\
         & Chronic gastritis & 618 & 77 & 77 & 772 \\
         & Erosion & 224 & 27 & 27 & 278 \\
         & Ulceration & 36 & 4 & 4 & 44 \\
        \hline
    \end{tabular}
\end{table*}

\section{Method}

\subsection{Dataset}
We employ a set of gastric endoscopic biopsy slides collected from 
The Catholic University of Korea Uijeongbu St. Mary's Hospital 
between 2014 and 2023. This dataset comprises 4,673 biopsy slides, each annotated with structured class labels, including both coarse-grained and fine-grained categories for detailed analysis. The details of this dataset and its hierarchical class labels are available in Table~\ref{tab:class_distribution}.

\subsection{Model Architecture}
The proposed method contains three key components: 1) Patch-level encoder $\mathcal{E}$; 2) Feature aggregator $\mathcal{A}$; 3) Hierarchical classifier $\mathcal{H}$. Each component is detailed below.

\subsubsection{Patch-level Encoder}
The patch-level encoder $\mathcal{E}$ receives a biopsy WSI, divides it into 512x512 image patches, and produces patch-level feature vectors. Specifically, for each image patch, we adopt UNI \cite{chen2024}, a pre-trained general-purpose model trained on over 100 million image patches via self-supervised learning, to produce a feature vector $\in \mathbb{R}^{1024}$.

\subsubsection{Feature Aggregator}
Given a set of patch-level feature vectors, the feature aggregator $\mathcal{A}$ produces a single representative feature vector for the entire slide.
Following CLAM \cite{lu2021}, we utilize attention-based pooling to effectively aggregate these patch-level features into a single feature vector $\in \mathbb{R}^{512}$. 

\subsubsection{Hierarchical Classifier}
The hierarchical classifier $\mathcal{H}$ simultaneously performs both coarse-grained classification and fine-grained classification through a structured pipeline with bidirectional feature integration, feature projection heads, and classification heads. 
Bidirectional feature integration receives an input feature vector from $\mathcal{A}$ and split it into two feature vectors: one designated as the coarse-grained feature vector $\mathbf{v}^c \in \mathbb{R}^{256}$ and the other as the fine-grained feature vector $\mathbf{v}^f \in \mathbb{R}^{256}$. Inspired by Chen et al. \cite{chang2021}, both feature vectors are then augmented with information from the other. To enrich the coarse-grained feature vector, we concatenate it with the fine-grained feature vector, while blocking the gradients from the fine-grained feature vector to prevent coarse-grained classification from being biased towards fine-grained representations. This procedure is given by $\mathbf{v}^{c'} = \mathbf{v}^c \circ G(\mathbf{v}^f) \in \mathbb{R}^{512}$ where $G$ is a gradient controller. Similarly, we augment the fine-grained feature vector as $\mathbf{v}^{f'} = \mathbf{v}^f \circ G(\mathbf{v}^c) \in \mathbb{R}^{512}$. In this manner, coarse-grained feature vectors retain fine-level details, while fine-grained feature vectors are equipped with high-level contextual information, without direct weight updates. 

Following bidirectional feature integration, both the augmented coarse-grained and fine-grained feature vectors are processed separately through a projection head and a classification head. The linear projection head $\mathcal{P}$ maps the augmented feature vectors into class-specific representations, generating one feature vector per class, as follows: $\mathbf{f}^{c} = \mathcal{P}_{c}(\mathbf{v}^{c'}) \in \mathbb{R}^{N_c \times 256}$ and $\mathbf{f}^{f} = \mathcal{P}_{f}(\mathbf{v}^{f'}) \in \mathbb{R}^{N_f \times 256}$. Then, the linear classification head takes the projected feature vectors and generates logits $\mathbf{o}^{c}$ and $\mathbf{o}^{f}$ corresponding to coarse-grained and fine-grained classification, respectively.

\subsection{Loss Functions}
The objective function for our method is given by $L=L_{CE}+L_{Con}+L_{Int}+L_{GCE}$ where $L_{CE}$, $L_{Con}$, $L_{Int}$, and $L_{GCE}$ denote cross-entropy loss, hierarchical consistency loss, intra- and inter-class distance loss, and group-wise cross-entropy loss, respectively. 
Each component plays a unique role in enforcing the hierarchical structure of class labels and improving discriminative ability of the learned representations, as illustrated in Figure~\ref{fig2}.

\subsubsection{Cross-Entropy Loss}
Cross-entropy loss $L_{CE}$ is defined as $L_{CE} = - \sum_{i} y_i \log p_i$ where \( y_i \) is the one-hot encoded ground truth for class $i$ and \( p_i \) is the predicted probability for class \( i \). $L_{CE}$ is independently applied to both coarse- and fine-grained classification tasks.
As a standard supervised loss, $L_{CE}$ serves as the foundation of the classification objective. However, it does not consider the hierarchical structure among classes, which motivates the inclusion of the additional loss terms introduced below.

\subsubsection{Hierarchical consistency Loss} 
$L_{Con}$ is designed to align the corresponding coarse-grained and fine-grained feature representations for improved hierarchical consistency in classification. Based on Jensen-Shannon Divergence (JSD) \cite{menendez1997}, $L_{Con}$ is defined as follows:
 \[
    L_{Con} = \frac{1}{2} D_{KL}(\mathbf{f}^{c*} || m) + \frac{1}{2} D_{KL}(\mathbf{f}^{f*} || m)
\]
where $\mathbf{f}^{c*}$ and $\mathbf{f}^{f*}$ denote the feature vectors corresponding to the highest logit values in $\mathbf{o}^{c}$ and $\mathbf{o}^{f}$, respectively, \( m = \frac{1}{2} ( \mathbf{f}^{c*} + \mathbf{f}^{f*} ) \) is the mixture distribution, and \( D_{KL} \) represents the Kullback-Leibler divergence. By enforcing the alignment between hierarchical class predictions, we encourage the model to produce similar feature representations for the corresponding coarse- and fine-grained classes. 
This alignment is crucial for maintaining semantic consistency across the class hierarchy. For example, if the coarse prediction is “cancer” but the fine prediction is “chronic gastritis”, this inconsistency can lead to diagnostic confusion.
$L_{Con}$ penalizes such semantic mismatches by minimizing divergence between each feature embeddings, thereby encouraging the network to produce more structured and interpretable predictions. 
Moreover, $L_{Con}$ operates on the learned feature vectors rather than logits, enabling more stable learning in multi-class environments.

\begin{table*}[htp]
    \centering
    \caption{Results of hierarchical classification on gastric endoscopic biopsy slides. The bold and underlined values indicate the best and second-best performance, respectively.}
    \label{tab:comparison_result}
    \begin{tabular}{llcccc}
        \hline
        \textbf{Model} & \textbf{Task} & \multicolumn{2}{c}{\textbf{Coarse-grained}} & \multicolumn{2}{c}{\textbf{Fine-grained}} \\
        \cline{3-6}
        & & \textbf{Accuracy(\%)} & \textbf{F1-macro} & \textbf{Accuracy(\%)} & \textbf{F1-macro} \\
        \hline
        MaxMIL & Coarse-grained & 80.99 & 0.8185 & - & - \\
         & Fine-grained & - & - & 65.23 & 0.4520 \\
         & Hierarchical & 78.19 & 0.7918 & 61.99 & 0.4026 \\
        MeanMIL & Coarse-grained & 82.07 & 0.8292 & - & - \\
         & Fine-grained & - & - & 66.31 & 0.4324 \\
         & Hierarchical & 83.15 & 0.8407 & 65.87 & 0.4135 \\
        CLAM-SB & Coarse-grained & \underline{84.88} & \underline{0.8565} & - & - \\
         & Fine-grained & - & - & 67.17 & 0.4650 \\
         & Hierarchical & 82.51 & 0.8348 & 66.52 & 0.4839 \\
        CLAM-MB & Coarse-grained & 82.94 & 0.8385 & - & - \\
         & Fine-grained & - & - & 67.60 & \underline{0.5046} \\
         & Hierarchical & 82.94 & 0.8419 & 68.47 & 0.4867 \\
        TransMIL & Coarse-grained & 81.21 & 0.8170 & - & - \\
         & Fine-grained & - & - & 61.34 & 0.3392 \\
         & Hierarchical & 79.70 & 0.8055 & 61.77 & 0.3604 \\
        S4MIL & Coarse-grained & 84.23 & 0.8496 & - & - \\
         & Fine-grained & - & - & 68.25 & 0.4320 \\
         & Hierarchical & 83.59 & 0.8454 & \underline{68.47} & 0.4880 \\
        Chang et al. \cite{chang2021} & Hierarchical & 83.80 & 0.8475 & 67.39 & 0.4871 \\
        HiClass (Ours) & Hierarchical & \textbf{85.10} & \textbf{0.8610} & \textbf{68.68} & \textbf{0.5220} \\
        \hline
    \end{tabular}
\end{table*}

\subsubsection{Intra- and Inter-class Distance Loss}
To enhance feature representations of $\mathbf{f}^{f}$, we propose $L_{Int}$ to enforce a structured separation between fine-grained classes based on their coarse-grained category. Specifically, $L_{Int}$ is designed to maximize the distance between fine-grained classes from the different coarse-grained category and minimize the distance between fine-grained classes from the same coarse-grained category. $L_{Int}$ is formulated as:
\[
    L_{Int} = \sum_{i+}D_{KL}(\mathbf{f}^{f}_{i*}||\mathbf{f}^{f}_{i+})
    + \sum_{i-}\max(0, \alpha - D_{KL}(\mathbf{f}^{f}_{i*}||\mathbf{f}^{f}_{i-}))
\]
where $i*$ is the ground truth fine-grained class index, $i+$ is the set of fine-grained class indices belonging to the same coarse-grained category as $y$, and $i-$ is the set of fine-grained indices belonging to a different coarse-grained category from $y$.
This objective promotes the formation of a hierarchical feature space, in which fine-grained class clusters naturally organize around their corresponding coarse-level categories.
These hierarchical constraints can be particularly beneficial for underrepresented fine-grained categories by providing additional contextual information from their coarse-level groupings. Even during fine-grained classification, the model can leverage coarse-level structure to improve inter-class separation and produce more informed, reliable predictions.

\subsubsection{Group-wise Cross-Entropy Loss}
To further improve fine-grained classification, we introduce $L_{GCE}$, which is given by:
\[
    L_{GCE} = - \sum_{k \in \mathcal{S}} y_k \log \frac{e^{\mathbf{o}^{f}_k}}{\sum_{j \in \mathcal{S}} e^{\mathbf{o}^{f}_j}}
\]
where \( \mathcal{S} \) represents the subset of fine-grained class indices belonging to the ground truth coarse-grained class category and $y$ is the one-hot encoded ground truth fine-grained class label. Unlike the standard cross-entropy loss, $L_{GCE}$ restricts the probability distribution within the fine-grained classes of the same coarse-grained category.
This effectively recalibrates fine-grained logits, leading to improved discrimination among fine-grained classes within the same coarse-grained category.
These constraints effectively sharpen class boundaries by reducing the number of competing logits during the softmax operation, thereby enabling more robust predictions. By shifting the optimization focus toward within-group discrimination, $L_{GCE}$ complements other loss functions that primarily aim to organize the feature space based on inter-group structure.
Moreover, this hierarchical mechanism reflects a natural diagnostic reasoning process, in which broad categories are identified first, followed by the determination of fine-grained subtypes.

\subsection{Implementation Details}
All experiments were conducted on an RTX A6000 GPU. All model were trained using a batch size of 1 for 20 epochs. The initial learning rate was set to 0.0001 and progressively decayed to 0.00001 using a cosine annealing scheduler. Optimization was performed using the Adam optimizer.

\section{Results}

\begin{table*}[htp]
    \centering
    \caption{Results of ablation experiments on HiClass. The bold values indicate the best performance.}
    \label{tab:ablation}
    \begin{tabular}{llccccccc}
        \hline
        \textbf{Task} & \textbf{Feature} & \multicolumn{3}{c}{\textbf{Loss Functions}} & \multicolumn{2}{c}{\textbf{Coarse-grained}} & \multicolumn{2}{c}{\textbf{Fine-grained}} \\
        \cline{3-9}
        & \textbf{Integration} & $\mathcal{L}_{Con}$ & $\mathcal{L}_{Int}$ & $\mathcal{L}_{GCE}$ & \textbf{Accuracy(\%)} & \textbf{F1-macro} & \textbf{Accuracy(\%)} & \textbf{F1-macro} \\
        \hline
        Coarse-grained & - & - & - & - & 82.72 & 0.8365 & - & - \\
        Fine-grained & - & - & - & - & - & - & 65.66 & 0.4569 \\ 
        \cline{1-9}
        Hierarchical & None & O & O & O & 83.80 & 0.8479 & 67.39 & 0.5052 \\
        & Fine→Coarse & O & O & O & 82.29 & 0.8319 & 65.66 & 0.4900 \\
        & Coarse→Fine & O & O & O & 82.07 & 0.8278 & 64.36 & 0.4878 \\
        \cline{2-9}
        & Bidirectional & X & X & X & 82.72 & 0.8362 & 65.66 & 0.5004 \\
        & Bidirectional & O & X & X & 83.15 & 0.8406 & 66.74 & 0.5124 \\
        & Bidirectional & X & O & X & 81.86 & 0.8290 & 65.01 & 0.4876 \\
        & Bidirectional & X & X & O & 82.29 & 0.8361 & 65.01 & 0.4897 \\
        & Bidirectional & X & O & O & 82.94 & 0.8420 & 63.93 & 0.4861 \\
        & Bidirectional & O & X & O & 82.72 & 0.8380 & 65.44 & 0.4725 \\
        & Bidirectional & O & O & X & 82.51 & 0.8345 & 68.03 & 0.4956 \\
        \cline{2-9}
        & Bidirectional & O & O & O & \textbf{85.10} & \textbf{0.8610} & \textbf{68.68} & \textbf{0.5220} \\
        \hline
    \end{tabular}
\end{table*}

\subsection{Hierarchical Classification Results}
We applied the proposed method to the diagnosis of gastric endoscopic biopsy slides. The performance of the method was separately assessed for coarse-grained classification and fine-grained classification, using two evaluation metrics: accuracy (Acc) and F1-macro score (F1). As shown in Table \ref{tab:comparison_result}, the proposed method achieved an accuracy of 85.10\% and F1-macro of 0.8610 for coarse-grained classification. 68.68\% accuracy and 0.5220 F1-macro were obtained for fine-grained classification. There is a performance gap between coarse-grained classification and fine-grained classification for both evaluation metrics. 
This is primarily ascribable to the increased difficulty and complexity of fine-grained classification in comparison to coarse-grained classification, where coarse-grained classification involves 4 classes, whereas fine-grained classification consists of 14 classes.
Moreover, fine-grained categories tend to exhibit greater visual and semantic similarity, which inherently increases the difficulty of accurate discrimination.

\subsection{Comparative Experiments}
We compared our method with various multiple instance learning (MIL) and hierarchical classification models designed for pathology image classification and natural image classification. These models include MaxMIL, MeanMIL, CLAM-SM and CLAM-MB \cite{lu2021}, TransMIL \cite{shao2021}, S4MIL \cite{fillioux2023}, and Chang et al. \cite{chang2021}. MaxMIL and MeanMIL use max pooling and mean pooling aggregation for slide-level classification, respectively. CLAM-SB and CLAM-MB \cite{lu2021} apply attention-based MIL with instance clustering to enhance feature separation. TransMIL \cite{shao2021} employs a transformer-based architecture to capture spatial and morphological relationships in WSIs. S4MIL \cite{fillioux2023} utilizes structured state space models (SSMs) to model long-range dependencies in patch sequences. Chang et al. \cite{chang2021} utilizes a hierarchical label structure for fine-grained and coarse-grained classification. Except for Chang et al. \cite{chang2021}, all these models were originally designed for flat classification. For a fair and comprehensive comparison, we separately trained these models for flat coarse-grained classification and flat fine-grained classification. Moreover, we modified their original architectures by incorporating dual classification heads for hierarchical classification, conducting simultaneous coarse-grained and fine-grained classification. 

Table~\ref{tab:comparison_result} depicts the results of gastric endoscopic biopsy slide classification using the proposed method and all competing models. The results show that HiClass attains the highest accuracy and macro-F1 at both the coarse-grained and fine-grained levels among all compared methods. However, competing models were inconsistent across classification tasks. For instance, CLAM-SB (flat classification) obtained the second-best performance for coarse-grained classification, whereas the best accuracy and F1-macro for fine-grained classification were attained by hierarchical S4MIL and flat CLAM-MB, respectively. Moreover, the effect of dual classification heads was disproportionate across competing models. It was generally beneficial for one classification task and was not able to consistently improve performance across both classification tasks. 
This suggests that simple architectural modifications may not be sufficient to effectively leverage hierarchical information. In contrast, the proposed method systematically incorporates hierarchical supervision through bidirectional feature integration and tailored loss functions, leading to more consistent and reliable performance improvements across both coarse- and fine-grained classification.
These observations suggest that the impact of hierarchical learning varies depending on the model architecture. Notably, Chang et al. \cite{chang2021}, specifically designed for hierarchical classification, was sub-optimal, further highlighting the superior performance of the proposed method.
This consistent performance across both classification levels demonstrates the robustness of our method to common challenges in pathology tasks, such as class imbalance and high inter-class similarity.

\subsection{Ablation Study}
To analyze the effect of different feature integration strategies and the loss functions on HiClass, we conducted a series of ablation experiments. For feature integration, we examined three scenarios: 1) None: No feature integration; 2) Fine$\rightarrow$Coarse: The fine-grained feature vector is concatenated to the gradient controlled coarse-grained feature vector; 3) Coarse$\rightarrow$Fine: the coarse-grained feature vector is concatenated to the fine-grained feature vector following the same procedure in 2) but in reverse order. To assess the impact of loss functions, we evaluated three loss functions ($L_{Con}$, $L_{KL-Div}$, and $L_{GCE}$) in various combinations, while $L_{CE}$ was used as a default loss function.

The results of ablation experiments are shown in Table~\ref{tab:ablation}. 
Unidirectional feature integrations (Fine$\rightarrow$Coarse and Coarse$\rightarrow$Fine) were less effective than the proposed bidirectional feature integration. It was not only inferior to the proposed method but also underperformed compared to no feature integration at all. 
In the investigation of the three loss functions, we found that no single loss function was dominant. The absence of each loss function resulted in a performance drop of 2.16\%$\sim$2.59\% accuracy and 0.019$\sim$0.0265 F1-macro for coarse-grained classification, and 0.65\%$\sim$4.75\% accuracy and 0.0264$\sim$0.495 F1-macro for fine-grained classification. 
The addition of $L_{Con}$ to $L_{CE}$ boosted the performance, but none of the other two loss functions alone yielded the better performance. These findings suggest that it is not an individual loss functions but the combined effect of all loss functions that contributes to the best performance obtained by the proposed method.
Each loss function contributes to a different aspect of hierarchical learning. The inter-group separation is supported by $L_{Int}$, intra-group discrimination is improved by $L_{GCE}$, and alignment between coarse and fine predictions is encouraged by $L_{Con}$. Together, these losses lead to more robust and well-structured representations.

\section{Conclusion}
In this study, we propose HiClass, a hierarchical classification framework designed to improve both coarse-grained and fine-grained pathology image classification. It enriches feature representations via bidirectional feature integration and effectively utilize them for hierarchical classification that simultaneously conducts coarse-grained classification and fine-grained classification. We also introduce tailored loss functions that facilitate improved hierarchical learning of diagnostic tasks in pathology images.
These loss functions contribute to structuring the feature space by enhancing inter-class separability, reinforcing intra-class discrimination within coarse-level groups, and ensuring coherence between hierarchical classification levels. This facilitates the learning of more semantically structured representations.
The experimental results demonstrate that the proposed method improves classification performance by incorporating class hierarchy into the model architecture and developing tailored learning strategies. The proposed approach is generic and applicable to a wide range of classification tasks and problems in pathology.

\section*{Acknowledgment} 
This work was supported by the National Research Foundation of Korea (NRF) (No. RS-2025-00558322), by the Ministry of Trade, Industry and Energy(MOTIE) and Korea Institute for Advancement of Technology (KIAT) through the International Cooperative R\&D program (P0022543), and the Catholic Medical Center Research Foundation made in the program year of 2025.

{
    \small
    \bibliographystyle{ieeenat_fullname}
    \bibliography{main}
}

\end{document}